\documentclass[letterpaper]{article} 
\usepackage[preprint]{aaai2027}
\usepackage[hyphens]{url}  
\usepackage{graphicx} 
\usepackage{natbib}  
\usepackage{caption} 
\usepackage{algorithm}
\usepackage{algorithmic}
\usepackage{amsmath}
\usepackage{amssymb}

\usepackage{newfloat}
\usepackage{listings}
\DeclareCaptionStyle{ruled}{labelfont=normalfont,labelsep=colon,strut=off} 
\floatstyle{ruled}
\newfloat{listing}{tb}{lst}{}
\floatname{listing}{Listing}
\usepackage{booktabs}
\usepackage{multirow}
\title{Beyond Geometry: Benchmarking and Consistency Reasoning for 3D Logical Anomaly Detection}
\author{Zhiqiang Qin\textsuperscript{1,2},
He Xie\textsuperscript{1,2}\thanks{Corresponding author: He Xie (\texttt{xiehe@hun.edu.cn}); Zhiqiang Qin: \texttt{qinzhiqiang@hun.edu.cn}.},
Junfei Yi\textsuperscript{1,2},
Yang Yang\textsuperscript{1,2},\\
Hao Wang\textsuperscript{1,2},
Yunkang Cao\textsuperscript{1,2},
Hui Zhang\textsuperscript{1,2},
Yaonan Wang\textsuperscript{1,2}}
\affiliations{\textsuperscript{1}School of Artificial Intelligence and Robotics, Hunan University\\
\textsuperscript{2}National Engineering Research Center of Robot Visual Perception and Control Technology, Hunan University}

\begin{document}

\maketitle

\begin{abstract}
Existing 3D industrial anomaly detection mainly targets local geometric deviations. In contrast, many industrial anomalies violate object-level design or assembly rules, which we define as 3D logical anomalies. To address these challenges, we introduce the Industrial Logical Anomaly Detection Dataset (\textbf{ILGAD}), the first scalable benchmark dedicated to logical anomalies in industrial point clouds. ILGAD contains 2,774 samples from 15 categories with point-level annotations and covers existence, specification, pose, and assembly-state errors. To detect such 3D logical anomalies, we propose a consistency reasoning framework that assesses whether local geometry, structure coverage, and spatial relations conform to the normal design. The framework detects geometric changes, unsupported expected structures, and abnormal local arrangements. Experiments on ILGAD, Anomaly-ShapeNet, and IEC3D demonstrate superior object-level detection and point-level localization, showing that the framework effectively detects logical anomalies and generalizes to conventional geometric defects.
\end{abstract}

\section{Introduction}

Industrial inspection often involves multi-component products and assemblies rather than only isolated parts. In these settings, correctness depends not only on surface quality but also on component presence, configuration, and inter-component relations~\cite{bergmann2022beyond}.
 An object may contain locally intact
components yet remain globally invalid because a required component
is absent, redundant, mismatched, misoriented, or incorrectly
assembled. This distinction leads to two fundamentally different inspection goals.
Conventional geometric anomaly detection asks where the observed
shape differs, whereas logical anomaly detection asks whether an
object satisfies its intended design and assembly rules. The need to
assess such rule violations has already motivated logical anomaly
detection in 2D industrial inspection. MVTec LOCO
AD~\cite{bergmann2022beyond} provides a 2D industrial benchmark
for logical anomaly detection, where logical anomalies violate
constraints on object presence or spatial arrangement. However,
image-based settings do not provide explicit 3D geometry for
evaluating component pose, spatial relations, and assembly states.
This limitation motivates the study of logical anomaly detection
directly in 3D point clouds, which provide explicit geometric cues for
assessing component configurations, spatial relations, and structural
validity.

\begin{figure}[t]
\centering
\includegraphics[width=\columnwidth]{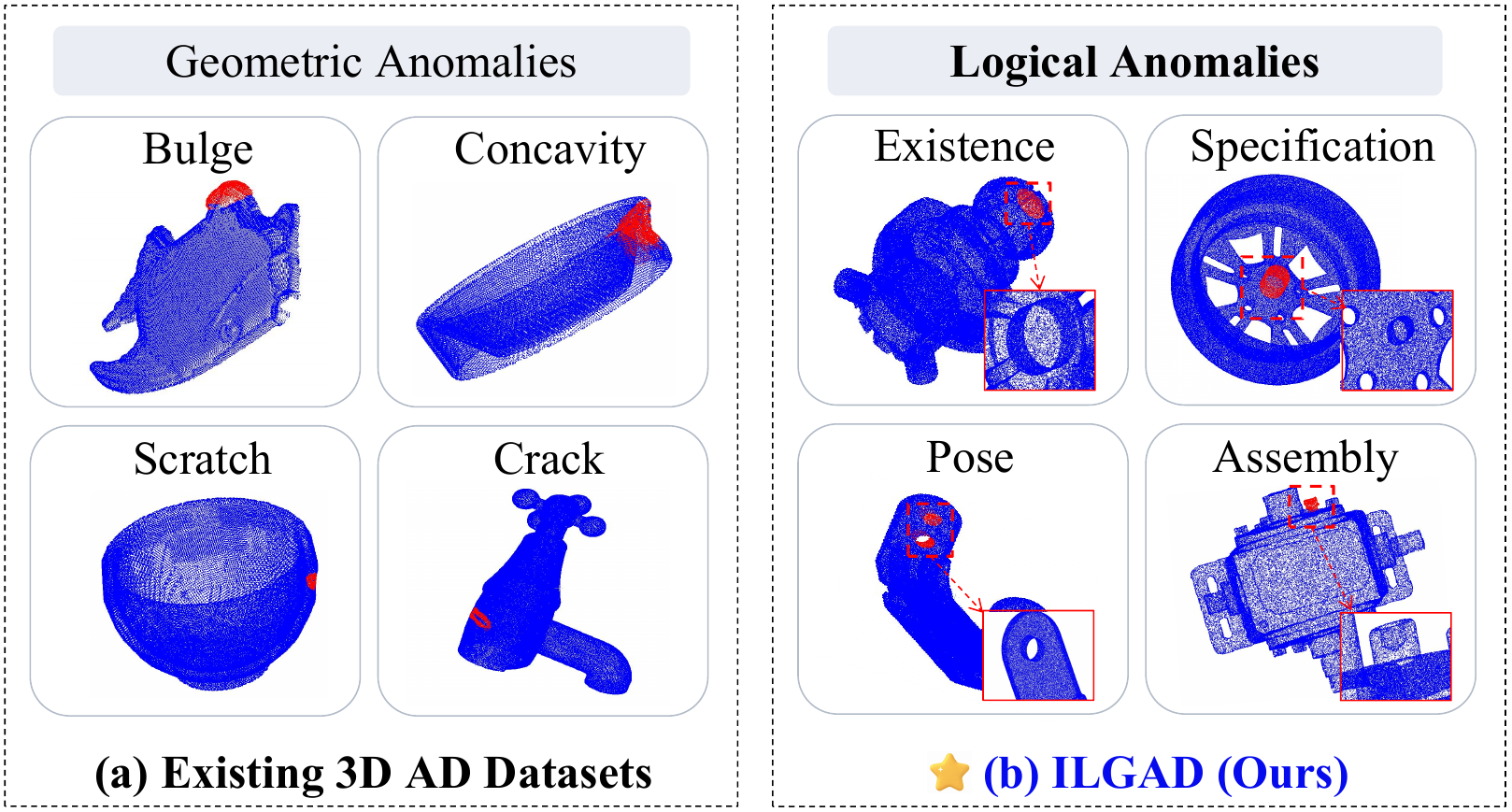}
\caption{Comparison between representative existing 3D anomaly detection datasets and ILGAD. Existing benchmarks mainly focus on geometric anomalies, whereas ILGAD covers existence, specification, pose, and assembly-state errors. Red markings indicate anomalies, and the insets show the corresponding normal structures.}
\label{fig:benchmark_comparison}
\end{figure}

Despite these advantages, most existing 3D anomaly detection methods still identify anomalies through local feature discrepancy, reconstruction error, registration residual, or point-wise deformation~\cite{horwitz2023back,cao2024cpmf,wang2023m3dm,zhou2024r3dad,liang2025ismp,ye2025po3ad}. These formulations are effective for scratches, dents, and local deformations because the anomalous evidence is directly present in the observed test geometry. However, they are less suited to logical violations that cannot be identified from local geometric evidence alone. A missing component contributes no test points to match, a wrongly posed component may retain normal local geometry, and an invalid assembly may consist entirely of individually normal parts. The central scientific question is therefore: \emph{Can a 3D inspection system determine whether an object is valid according to its intended structure and assembly rules, rather than merely detecting geometric deviations?}

Answering this question requires not only a suitable reasoning
framework but also a benchmark that explicitly represents
object-level structural and assembly violations. Existing benchmarks
cover real scanned objects, synthetic shape anomalies, multi-sensor
inspection, and subtle local defects~\cite{bergmann2022mvtec,
liu2024real3dad,li2024anomalyshapenet,li2024mulsenad,
cheng2025minishift, zhang2026unification}. However, as illustrated by the representative
examples in Figure~\ref{fig:benchmark_comparison}(a), these benchmarks
mainly focus on visible geometric anomalies and do not explicitly
organize anomalies around object-level design and assembly rules.
Consequently, they cannot fully evaluate whether a detector can move
beyond local shape comparison and determine whether an industrial
object is structurally valid. To address this limitation, we construct the Industrial Logical
Anomaly Detection Dataset (ILGAD). As shown in
Figure~\ref{fig:benchmark_comparison}(b), ILGAD defines four types of
logical anomalies, including existence, specification, pose, and
assembly-state errors. It contains 2,774 point clouds from 15
industrial categories and provides rule-guided CAD anomaly modeling
and point-level annotations, together with a small number of geometric
defects for mixed-anomaly evaluation.

Logical anomalies can be identified by examining whether an object remains consistent with its normal design at three complementary levels: local geometry, structure coverage, and spatial relations. We therefore ask three questions. Does each visible local structure match its expected geometry (\textbf{Q1})? Is every expected structure supported by the test observation (\textbf{Q2})? Are the normal spatial relations among neighboring structures preserved (\textbf{Q3})? Based on this formulation, we propose a logical consistency reasoning framework. It addresses \textbf{Q1}, \textbf{Q2}, and \textbf{Q3} through local geometric consistency, structure coverage consistency, and spatial relation consistency, respectively. Violations of these consistencies reveal specification errors, existence errors, and pose or assembly-state errors. The framework produces both point-level anomaly maps and object-level predictions. Our contributions are summarized below.
\begin{itemize}
    \item We formulate 3D logical anomaly detection as structural validity assessment beyond local geometric deviation and introduce ILGAD, the first scalable industrial point cloud benchmark dedicated to logical anomalies, with 15 categories, 2,774 samples, and point-level annotations.
    \item We propose a consistency reasoning framework that moves beyond local
shape comparison by jointly reasoning about visible geometry,
expected-structure coverage, and spatial relations, thereby detecting
visible defects, missing structures, and invalid local arrangements.
    \item Extensive experiments on ILGAD, Anomaly-ShapeNet, and IEC3D demonstrate superior object-level detection and point-level localization, validating the effectiveness and generalization of the proposed framework.
\end{itemize}

\begin{table}[t]
\centering
{\small

\begin{tabular}{@{}lcccc@{}}
\toprule
Dataset & Logical & Rule-guided & GT & Asm. Obj. \\
\midrule
MVTec 3D-AD
& $\times$ & $\times$ & \checkmark & $\times$ \\
Real3D-AD
& $\times$ & $\times$ & \checkmark & $\times$ \\
Anomaly-ShapeNet
& $\times$ & $\times$ & \checkmark & $\times$ \\
IEC3D
& $\times$ & $\times$ & \checkmark & $\times$ \\
ILGAD
& \checkmark & \checkmark & \checkmark & \checkmark \\
\bottomrule
\end{tabular}
}
\caption{Comparison of representative 3D anomaly detection
datasets. Rule-guided indicates that anomalous CAD models are manually
constructed by violating predefined structural or assembly requirements. Asm. Obj. indicates whether the dataset contains multi-component
assembled objects.}
\label{tab:dataset_comparison}
\end{table}

\section{ILGAD Benchmark}

\subsection{Task Definition}
Let \(O=\{p_i\}_{i=1}^{N}\) denote an observed industrial point
cloud and \(R\) a normal reference from the same category. A logical
anomaly occurs when \(O\) violates an object-level design or assembly
constraint represented by \(R\), resulting in an existence,
specification, pose, or assembly-state error. The task includes object-level detection and point-level localization.
The object-level label \(y_{\mathrm{obj}}\in\{0,1\}\) indicates whether
\(O\) is anomalous, while the point-level labels
\(\mathbf{y}_{\mathrm{pt}}=\{y_i\}_{i=1}^{N}\), where
\(y_i\in\{0,1\}\), identify visible evidence of the violation. Table~\ref{tab:dataset_comparison} compares ILGAD with representative
3D anomaly detection datasets, including MVTec 3D-AD
~\cite{bergmann2022mvtec}, Real3D-AD~\cite{liu2024real3dad},
Anomaly-ShapeNet~\cite{li2024anomalyshapenet}, and IEC3D
~\cite{guo2025iec3dad}. Unlike existing benchmarks that mainly focus
on visible geometric defects, ILGAD explicitly models logical
violations of industrial design and assembly rules.

\subsection{Dataset Construction}

ILGAD is constructed in four stages. First, professional industrial
design engineers define 15 generic industrial component categories
containing common structures such as holes, grooves, repeated elements,
and assembled subcomponents. Second, defect-free CAD templates are
created in SolidWorks, from which normal point clouds are generated
with varying sampling densities and global poses. Third, anomalous
variants are manually created by violating explicit design and assembly
rules, ensuring meaningful logical defects rather than arbitrary
geometric perturbations. Finally, all CAD models are converted into point clouds and manually
annotated at the point level in CloudCompare. For missing-component
anomalies, the removed structure has no corresponding points in the
test cloud. We therefore annotate the visible contact or boundary
points adjacent to the missing location as point-level ground truth.
These labels capture the observable evidence of structural absence
while ensuring that the predictions and ground-truth annotations are
defined on the same test point set. Logical anomalies are categorized as existence, specification, pose,
and assembly-state errors. They respectively describe missing or
redundant components, incorrect structural parameters, incorrect
component positions or orientations, and invalid assembly states.
ILGAD also includes geometric defects for conventional and mixed-anomaly
evaluation. Since a test sample may contain multiple logical violations
or both logical and geometric defects, the taxonomy describes violated
rules rather than mutually exclusive sample classes. Following the
standard industrial anomaly detection protocol, the training split
contains only defect-free samples, while the test split contains normal
and anomalous samples with point-level annotations. No ILGAD anomaly is
used for training or model selection.
\begin{figure*}[t]
\centering
\includegraphics[width=\textwidth]{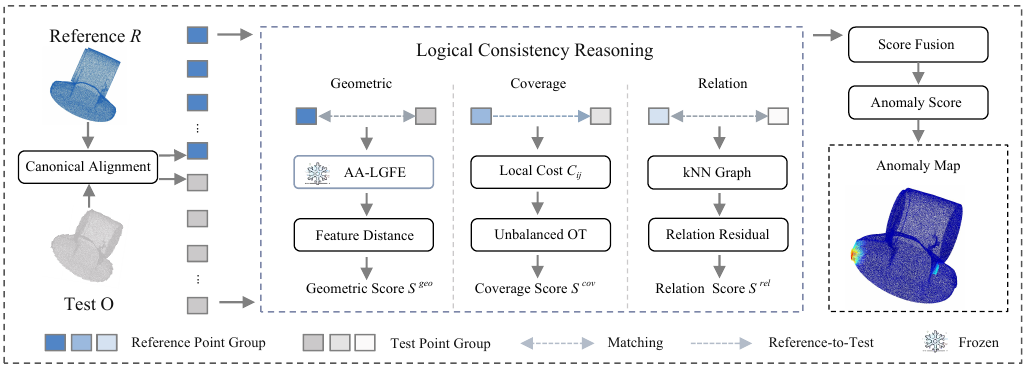}
\caption{Overview of the Consistency Reasoning Framework.
After canonical reference alignment, three parallel components
evaluate local geometric consistency, structure coverage consistency,
and spatial relation consistency. The resulting scores are propagated
to visible test points and fused to produce a point-level anomaly map
and an object-level anomaly score.}
\label{fig:framework}
\end{figure*}

\subsection{Dataset Statistics}

ILGAD contains 2,774 samples from 15 industrial component categories
in total, including 1,602 normal samples and 1,172 anomalous samples. Among the 1,172 anomalous samples, 908 contain at least one logical violation, including 248 samples with mixed logical and geometric defects, while the remaining 264 samples contain geometric defects only. The categories are Cylindrical Bracket (CLB), Bearing Connector Assembly (BC), Wheel Hub (WH), Box Connector Assembly (BCA), Curved Bracket (CDB), Rectangular Cover (RC), Pipe Bracket (PB), Circular Connector (CC), Mechanical Mounting Base (MMB), Automotive End Cap (AEC), Valve Component (VC), Elbow Joint (EJ), Rail Bracket Assembly (RBA), Rectangular Mounting Plate (RMP), and Oval Mounting Flange (OMF). Each point cloud contains approximately 300K--500K points after preprocessing. Detailed category-wise
train--test splits, non-exclusive anomaly-type distributions, and
operational definitions for anomaly generation and annotation are
provided in the supplementary material for completeness and
reproducibility.

\section{Consistency Reasoning Framework}

\subsection{Overview}
Given the test observation \(O\) and the normal reference \(R\) from the same category, the framework evaluates three complementary forms of consistency. Local geometric consistency measures visible patch-level changes. Structure coverage consistency evaluates whether each reference structure is supported by the observation, including structures that
may be absent from $O$. Spatial relation consistency measures changes
in the relative arrangement of neighboring local structures. As shown in Figure~\ref{fig:framework}, the reference and test
point clouds are first aligned to a shared canonical frame. In the
local geometric consistency branch (\textbf{Q1}), matched reference
and test patches are encoded by a shared Anomaly-Aware Local
Geometric Feature Encoder (AA-LGFE). The structure coverage
consistency branch (\textbf{Q2}) operates on aligned point coordinates
to verify whether expected reference structures are supported by the
observation, while the spatial relation consistency branch
(\textbf{Q3}) operates on neighborhood structures to assess abnormal
local arrangements. The resulting geometric, coverage, and relation
scores are then propagated to the visible test points, normalized, and
fused for point-level localization and object-level detection.

\subsection{Canonical Reference Alignment}
To match component-level structure, we canonicalize the normal reference and test point cloud into a shared coordinate frame. Rather than relying on expensive unconstrained global registration, we estimate a stable canonical basis by anchor-set voting and then apply small-range ICP refinement.

For each point cloud
\(\mathcal{X}=\{x_i\}_{i=1}^{N}\), we first robustly center
the points as \(\bar{x}_i=x_i-c\). We then compute the radial
distance \(r_i=\|\bar{x}_i\|_2\) and select outer and middle
anchor sets according to radial-distance quantiles. The outer-anchor set $\mathcal{A}_{\mathrm{out}}$ captures the
global object extension and is used to estimate the first axis, while the
middle-anchor set $\mathcal{A}_{\mathrm{mid}}$ is less sensitive to boundary
noise and extreme outliers and is used to estimate the second axis. The first
axis is obtained from the weighted direction-voting matrix:
\begin{equation}
A=
\sum_{i\in\mathcal{A}_{\mathrm{out}}}
w_i\hat{x}_i\hat{x}_i^{\top},
\qquad
\hat{x}_i=
\frac{\bar{x}_i}{\|\bar{x}_i\|_2+\epsilon}.
\end{equation}
Here, $w_i$ is a normalized robust density-aware weight computed from the
local $k$NN spacing:
\begin{equation}
\delta_i=
\frac{1}{k}
\sum_{j\in\mathcal{N}_k(i)}
\|\bar{x}_i-\bar{x}_j\|_2,
\end{equation}
where $\mathcal{N}_k(i)$ is obtained from the full centered point cloud.
Specifically, $w_i$ combines a density-compensation factor derived from
$\delta_i$ and a Huber-type robust reliability factor
\cite{huber1964robust}. The density-compensation factor reduces the influence
of non-uniform sampling, while the robust factor suppresses isolated outliers
and unstable boundary points. The first canonical axis $e_1$ is the
eigenvector corresponding to the largest eigenvalue of $A$.

To estimate the second axis, each middle center is first projected onto
the subspace orthogonal to $e_1$ as
$z_i=\bar{x}_i-(\bar{x}_i^{\top}e_1)e_1$. A weighted covariance matrix is
then computed from the projected middle centers. Its dominant eigenvector
is used to obtain the second axis $e_2$ after removing any remaining
component along $e_1$. The third axis is obtained by
$e_3=e_1\times e_2$. The basis $[e_1,e_2,e_3]$ defines the canonical
coordinate frame. To resolve axis-sign ambiguity, we evaluate the four right-handed
sign configurations and select the one with the smallest symmetric
Chamfer distance to the category-level normal reference. A small-range
ICP refinement is then applied between the test cloud and the reference
template to improve local alignment.

\subsection{Anomaly-Aware Feature Encoder}

Dense patch-wise comparison requires local descriptors that are
both anomaly-sensitive and computationally lightweight. We therefore introduce AA-LGFE, a lightweight network pretrained
with source-domain point-level normal-versus-anomalous supervision. The anomaly-aware objectives promote
compact normal representations and improve their separation from
anomalous local structures.

AA-LGFE consists of a shared stem, a stable-geometry branch,
a multi-scale geometry branch, and a gated fusion module.
The stable-geometry branch uses stacked depthwise-separable
blocks to progressively model regular local patterns over
distance-ordered neighbors. In contrast, the multi-scale
geometry branch employs parallel paths with kernel sizes
1, 3, and 5 to capture geometric structures under different
receptive fields. For a patch centered at \(q_j\), we construct
\(U_j\in\mathbb{R}^{K\times7}\) from its \(K\) nearest neighbors,
where each point contains normalized relative coordinates, a surface
normal, and normalized distance to the patch center. The shared stem
maps the patch into feature space:
\begin{equation}
H_j=\phi_2\bigl(\phi_1(U_j)\bigr),
\end{equation}
where \(\phi_1\) and \(\phi_2\) are \(1\times1\)
Conv-BN-ReLU layers. The two branches and gated fusion are given by
\begin{equation}
\begin{gathered}
H_{s,j}=D_s(H_j), \qquad H_{v,j}=D_v(H_j),\\
[\alpha_{s,j},\alpha_{v,j}]
=\operatorname{Softmax}\bigl(\operatorname{MLP}(g_j)\bigr),\\
H_{f,j}
=\alpha_{s,j}H_{s,j}+\alpha_{v,j}H_{v,j},
\end{gathered}
\end{equation}
where \(g_j\) is obtained by pooling the branch features.
The fused feature \(H_{f,j}\) is aggregated and projected into the
normalized patch descriptor \(f_j\).

A source patch is labeled anomalous when its anomalous-point
fraction is at least $0.05$. During pretraining, a temporary binary patch classifier and
feature-space objectives encourage compact representations within
the normal and anomalous classes and increase their separation.
The classifier is then discarded, and the frozen AA-LGFE is shared
by the reference and test branches for feature extraction.
Additional feature-only comparisons with FPFH and a frozen
PointMAE encoder are provided in the supplementary material.

\begin{table*}[!t]
\centering
{\small
\setlength{\tabcolsep}{1mm}

\begin{tabular}{lccccccccc}
\toprule
Category & PC-FPFH & PC-MAE & BTF-Raw & BTF-FPFH & Reg3D-AD & PO3AD & Template3D & Simple3D & Ours \tabularnewline
&  CVPR'22 &  CVPR'22 &  CVPRW'23 &  CVPRW'23 &  NeurIPS'23 &  CVPR'25 &  IJCAI'25 &  AAAI'26 &   \tabularnewline
\midrule
CLB & \underline{90.61}/72.49 & 42.07/55.69 & 47.20/47.30 & 58.00/70.20 & 83.85/64.81 & 57.34/65.48 & 72.83/\underline{95.75} & 65.00/68.90 & \textbf{100.00}/\textbf{99.25} \tabularnewline
BC & \textbf{82.05}/78.26 & 61.05/60.03 & 54.00/56.00 & 52.50/58.00 & \underline{78.12}/81.61 & 47.45/65.89 & 52.30/\underline{90.73} & 62.50/63.90 & 77.95/\textbf{94.99} \tabularnewline
WH & 65.43/77.15 & 47.97/64.16 & 55.60/58.00 & 46.20/49.30 & 65.85/83.41 & \underline{67.95}/79.61 & \textbf{68.05}/\underline{84.82} & 58.00/54.30 & 61.52/\textbf{86.68} \tabularnewline
BCA & \underline{81.55}/82.47 & 50.53/71.15 & 46.90/61.10 & 47.50/60.70 & 67.53/73.44 & 60.68/74.84 & 53.25/\underline{95.02} & 78.00/64.60 & \textbf{99.72}/\textbf{98.95} \tabularnewline
CDB & 88.49/79.14 & 72.75/76.05 & 56.30/45.60 & 53.60/63.80 & 81.81/84.12 & 64.31/67.67 & \underline{91.11}/\underline{93.18} & 73.30/55.60 & \textbf{99.78}/\textbf{94.98} \tabularnewline
RC & \underline{77.29}/62.98 & 51.24/55.07 & 50.50/48.80 & 47.00/64.20 & 75.51/66.76 & 62.96/72.18 & 63.71/\underline{78.24} & 53.00/72.40 & \textbf{100.00}/\textbf{98.38} \tabularnewline
PB & 94.51/78.63 & 55.03/51.89 & 61.00/59.10 & 52.70/61.90 & 86.97/77.86 & 57.37/61.18 & \underline{95.29}/\underline{93.52} & 67.90/63.60 & \textbf{100.00}/\textbf{98.77} \tabularnewline
CC & \underline{67.07}/69.19 & 48.25/52.34 & 51.20/57.50 & 55.90/64.40 & 60.47/68.07 & 55.53/56.17 & \textbf{71.44}/\underline{83.03} & 63.40/70.70 & 59.66/\textbf{83.45} \tabularnewline
MMB & 77.03/72.16 & 58.57/60.24 & 60.70/47.80 & 48.90/61.30 & \underline{83.24}/68.14 & 57.42/60.48 & 77.73/\underline{86.36} & 64.10/67.00 & \textbf{100.00}/\textbf{99.37} \tabularnewline
AEC & 51.95/68.73 & 52.44/67.01 & 34.80/41.20 & 45.50/51.20 & 54.41/83.17 & 48.49/58.58 & \underline{78.79}/\underline{93.99} & 42.90/54.70 & \textbf{92.84}/\textbf{96.20} \tabularnewline
VC & \underline{73.30}/72.18 & 49.61/53.60 & 44.20/52.10 & 49.20/55.40 & 65.80/69.68 & 45.38/49.66 & 66.00/\underline{81.34} & 43.50/62.00 & \textbf{97.81}/\textbf{98.40} \tabularnewline
EJ & 61.15/73.01 & 52.19/53.67 & 42.20/48.60 & 49.70/39.30 & 52.01/54.50 & 53.20/44.53 & \underline{84.11}/\underline{73.84} & 64.10/43.00 & \textbf{97.10}/\textbf{97.88} \tabularnewline
RBA & 74.26/62.57 & 72.30/62.07 & 66.30/67.20 & 67.20/66.40 & 62.27/75.86 & 60.13/71.52 & \underline{90.66}/\underline{98.66} & 56.20/75.20 & \textbf{100.00}/\textbf{99.77} \tabularnewline
RMP & 63.82/59.18 & 55.50/57.72 & 50.10/69.50 & 53.00/63.10 & 59.56/75.64 & 67.13/78.79 & 91.26/\underline{96.81} & \textbf{100.00}/74.30 & \underline{96.75}/\textbf{97.48} \tabularnewline
OMF & 78.01/61.34 & 57.83/69.27 & 56.90/70.60 & 60.20/64.40 & 62.87/74.26 & 82.84/90.54 & 82.10/\underline{97.06} & \textbf{99.00}/74.40 & \underline{95.39}/\textbf{99.68} \tabularnewline
\midrule
Mean & 75.10/71.30 & 55.16/60.66 & 51.86/55.36 & 52.47/59.57 & 69.35/73.42 & 59.21/66.47 & \underline{75.91}/\underline{89.49} & 66.10/64.30 & \textbf{91.90}/\textbf{96.28} \tabularnewline
\bottomrule
\end{tabular}
}
\caption{Quantitative results on ILGAD. The results are reported as O-ROC\%/P-ROC\%. The best performance is in \textbf{bold}, and the second best is \underline{underlined}.}
\label{tab:ilgad_results}
\end{table*}

\subsection{Logical Consistency Reasoning}

\paragraph{Local Geometric Consistency:}
Local geometric consistency compares matched
test and reference patches to identify local structural deviations. For each test center \(q_i\), we find its nearest reference center \(\hat{q}_i\) in the canonical frame. The matched test and reference patches are encoded by the shared AA-LGFE to obtain descriptors \(f_i^O\) and \(f_i^R\), respectively. Their geometric inconsistency can be described as:
\begin{equation}
s_i^{\mathrm{geo}}
=
\|f_i^O-f_i^R\|_2.
\end{equation}
This score captures visible geometric and specification changes but cannot directly represent a component that is absent from the test cloud.

\paragraph{Structure Coverage Consistency:}

We perform reference-side verification to
determine whether each expected reference structure is supported by
the test observation. A missing component has no test-side patch to
match. Therefore, the verification direction is reversed from the
reference to the test observation. For each reference anchor \(a_l\), we construct an expected patch \(R_l\) and an observed candidate patch \(O_l\) at the same canonical location.

For each reference point \(r_i\in R_l\) and observed candidate point \(p_j\in O_l\), we compute the squared Euclidean distance:
\begin{equation}
D_{ij}=\|r_i-p_j\|_2^2 .
\end{equation}
To make the comparison adaptive to local object scale, the distance matrix is normalized by a reference-side local scale:
\begin{equation}
C_{ij}=\frac{D_{ij}}{\sigma_l+\epsilon},
\end{equation}
where \(\sigma_l\) is the median squared distance from reference patch points to the anchor \(a_l\). This reference-side normalization prevents missing structures from being suppressed when all nearby test points are far from the expected normal structure.

The normalized pairwise costs form the transport cost matrix
$C=[C_{ij}]$. Here, $|R_l|$ and $|O_l|$ denote the numbers of
points in the local reference set $R_l$ and the corresponding
observed candidate set $O_l$, respectively. We assign uniform
masses to individual reference and observed points as
$\mu_i=1/|R_l|$ and $\nu_j=1/|O_l|$. The corresponding mass
vectors are
$\boldsymbol{\mu}=(\mu_i)_{i=1}^{|R_l|}$ and
$\boldsymbol{\nu}=(\nu_j)_{j=1}^{|O_l|}$.
Using $C$, $\boldsymbol{\mu}$, and $\boldsymbol{\nu}$, we compute
an unbalanced transport plan $T$ between $R_l$ and
$O_l$~\cite{cuturi2013sinkhorn,chizat2018scaling}. The transported mass measures how much expected reference structure is supported by the observation, while unmatched reference mass indicates missing structural evidence. For each reference point, the missing mass is defined as:
\begin{equation}
m_i=\max\left(0,\mu_i-\sum_j T_{ij}\right),
\end{equation}
where \(\mu_i\) denotes the reference-side mass. The coverage-gap score at anchor \(a_l\) can be described as:
\begin{equation}
s_l^{cov}=\sum_i m_i .
\end{equation}
A high \(s_l^{\mathrm{cov}}\) indicates that the expected structure around \(a_l\) is insufficiently supported by the observation.

\paragraph{Spatial Relation Consistency:}
Spatial relation consistency evaluates whether
neighboring local structures preserve the relative lengths and
directions observed in the normal reference. Some logical anomalies
retain plausible local geometry while violating these spatial
relations. To capture such cases, spatial relation consistency matches local graph relations between the test cloud and the corresponding normal structure. We build a KNN graph over sampled test centers. For each test edge \((q_i,q_j)\), we use the matched reference centers \(\hat{q}_i\) and \(\hat{q}_j\) to define:
\begin{equation}
d_{ij}^{O}=q_j-q_i,
\qquad
d_{ij}^{R}=\hat{q}_j-\hat{q}_i.
\end{equation}
The relation inconsistency score is computed by averaging the length and direction residuals over local neighbors:
\begin{equation}
s_i^{rel}=
\frac{1}{|\mathcal{N}(i)|}
\sum_{j\in\mathcal{N}(i)}
\left(\ell_{ij}+\eta o_{ij}\right),
\end{equation}

where \(\ell_{ij}=\left|\log\frac{\|d^O_{ij}\|_2+\epsilon}{\|d^R_{ij}\|_2+\epsilon}\right|\) measures relative length change, and \(o_{ij}=1-\frac{(d^O_{ij})^\top d^R_{ij}} {\max(\|d_{ij}^{O}\|_2,\epsilon_r)
 \max(\|d_{ij}^{R}\|_2,\epsilon_r)}\) measures direction inconsistency. We set \(\eta=0.5\).

\subsection{Consistency Fusion}
The local geometric and spatial relation scores are initially
computed at sampled test centers, whereas the structural coverage
scores are defined at reference anchors. We interpolate the
geometric and relation scores from the sampled test centers to
visible test points using local kNN interpolation. The coverage
scores are propagated from the reference anchors to nearby visible
test points. After normalization, the resulting aligned
point-level score maps are denoted by:
$\bar{s}_{i}^{\mathrm{geo}}$,
$\bar{s}_{i}^{\mathrm{cov}}$, and
$\bar{s}_{i}^{\mathrm{rel}}$.
The fused point-level anomaly score is computed as
\begin{equation}
S_i =
\alpha_{\mathrm{geo}}\bar{s}_{i}^{\mathrm{geo}}
+\alpha_{\mathrm{cov}}\bar{s}_{i}^{\mathrm{cov}}
+\alpha_{\mathrm{rel}}\bar{s}_{i}^{\mathrm{rel}},
\end{equation}
where
$\alpha_{\mathrm{geo}}
=\alpha_{\mathrm{cov}}
=\alpha_{\mathrm{rel}}=1/3$.
The object-level invalidity score is obtained by top-$0.1\%$
pooling:
\begin{equation}
S_{\mathrm{obj}}
=
\frac{1}{|\Omega|}
\sum_{i\in\Omega}S_i,
\end{equation}
where $\Omega$ contains the top $0.1\%$ of test points
ranked by $S_i$.

\section{Experiments}
\subsection{Experimental Setups}

\paragraph{Datasets:} We evaluate on ILGAD, Anomaly-ShapeNet~\cite{li2024anomalyshapenet}, and IEC3D~\cite{guo2025iec3dad}. ILGAD evaluates logical and mixed anomalies, while Anomaly-ShapeNet and IEC3D assess generalization to synthetic and real-scanned geometric defects.

\paragraph{Implementation Details:}
AA-LGFE is pretrained once on Real3D-AD using source-domain point-level anomaly annotations and is then frozen for all target datasets. Therefore, no target-domain anomaly is used, while the encoder benefits from external anomaly supervision. At inference, the geometric and relational branches sample 4,096 test centers by farthest-point sampling and extract 128-dimensional descriptors from 64-neighbor patches. The coverage branch samples 2,048 reference anchors, and the relational graph uses a $k=8$ nearest-neighbor query. The three normalized score maps are fused with equal weights. Additional details of AA-LGFE pretraining, point-cloud
canonicalization, unbalanced transport, and hyperparameter settings
are provided in the supplementary material.

\paragraph{Metrics:}
We report object-wise AUROC (O-ROC) and point-wise AUROC (P-ROC) as the primary metrics for object-level anomaly detection and point-level anomaly localization. Category-level results and additional metrics, including O-AP, P-AP, P-Best F1, and P-IoU@F1$^*$, are reported in
the supplementary material.

\paragraph{Comparison Methods:}
We compare our method with representative classical and recent
3D anomaly detection methods on ILGAD,
including feature- and memory-bank-based methods, i.e., 
BTF-Raw/BTF-FPFH~\cite{horwitz2023back}, 
PC-FPFH~\cite{roth2022patchcore,rusu2009fpfh}, 
and PC-MAE~\cite{roth2022patchcore,pang2022pointmae}; 
registration- or prototype-based methods, i.e., 
Reg3D-AD~\cite{liu2024real3dad} and Simple3D~\cite{cheng2025minishift}; 
the regression-based method PO3AD~\cite{ye2025po3ad}; 
and the template-guided method Template3D~\cite{liu2025template3dad}. 
For feature variants, ``-FPFH'' and ``-MAE'' denote FPFH~\cite{rusu2009fpfh} 
and PointMAE~\cite{pang2022pointmae} features, respectively. 
For Anomaly-ShapeNet and IEC3D, we further include representative results reported 
on the corresponding benchmarks, including 
M3DM~\cite{wang2023m3dm}, 
CPMF~\cite{cao2024cpmf}, 
IMRNet~\cite{li2024anomalyshapenet}, 
R3D-AD~\cite{zhou2024r3dad}, 
ISMP~\cite{liang2025ismp}, 
PO3AD~\cite{ye2025po3ad}, 
AF3AD~\cite{balapour2026af3ad}, 
MC3D-AD~\cite{cheng2025mc3dad}, 
GMANet~\cite{guo2025iec3dad}, 
Point-Patch~\cite{kang2026pointpatchfusion}, 
and SeDiR~\cite{kim2026sedir}.

\begin{figure*}[t]
\centering
\includegraphics[width=0.98\textwidth]{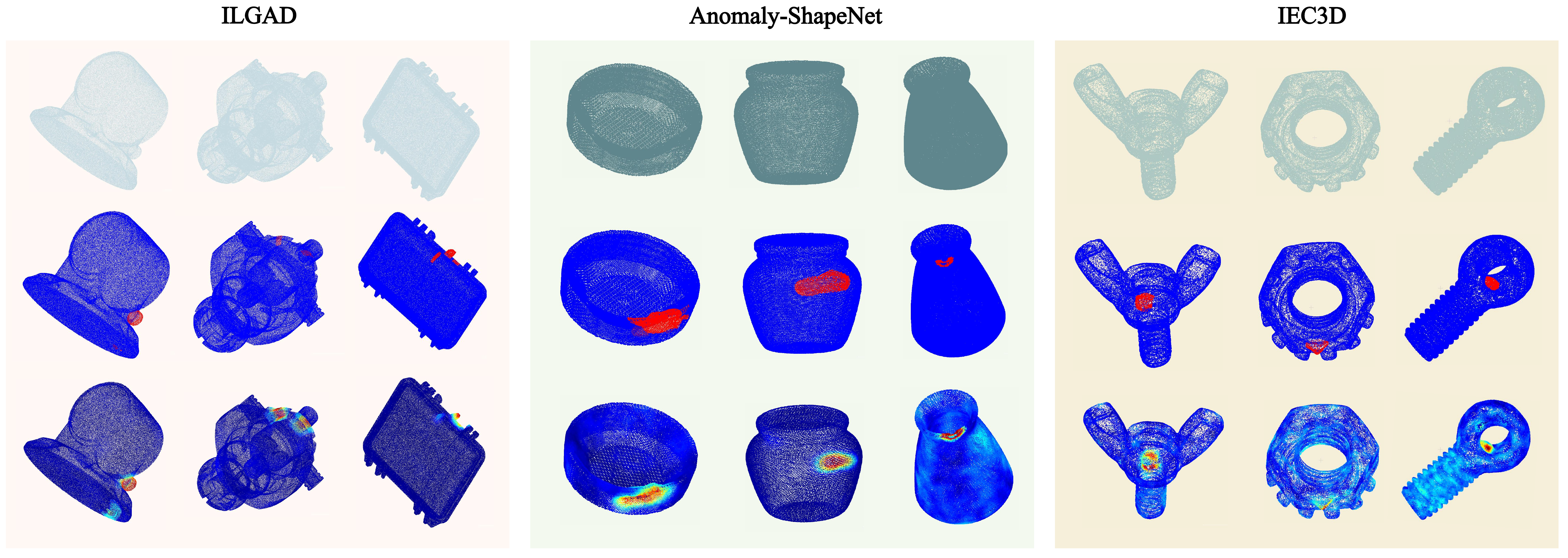}
\caption{Qualitative anomaly localization results on ILGAD, Anomaly-ShapeNet, and IEC3D. For each dataset, the top row shows the input point clouds, the middle row shows the point-level ground truth, and the bottom row shows the predicted anomaly maps. Red regions indicate anomalous areas or high anomaly responses.}
\label{fig:qualitative}
\end{figure*}

\subsection{Benchmarking Results on ILGAD}

Table~\ref{tab:ilgad_results} reports category-wise results on ILGAD. Our framework achieves 91.90\% O-ROC and 96.28\% P-ROC, surpassing the strongest mean baseline, Template3D, by 15.99 and 6.79 percentage points, respectively. This gain shows the benefit of jointly evaluating visible geometry, structure coverage, and local spatial relations. Most baselines identify visible local changes but do not explicitly verify unsupported expected structures or abnormal relations between neighboring local structures. The high P-ROC further indicates that reference-side violations can be effectively projected onto observable boundary or contact regions.

\subsection{Generalization to Geometric Anomalies}
\paragraph{Anomaly-ShapeNet:}
The quantitative comparisons are shown in Table~\ref{tab:anomaly_shapenet_results}. 
Our method achieves the best performance on both object-level and point-level metrics, 
with 94.39\% O-ROC and 97.33\% P-ROC. Compared with the second-best results, our method improves O-ROC over SeDiR by 1.09 percentage points and P-ROC over AF3AD by 4.83 percentage points.
These results show that although our framework is designed for logical anomaly detection, 
it also generalizes well to conventional geometric defect detection. 
The gain in P-ROC further indicates the proposed framework can provide accurate 
localization for local shape defects.

\paragraph{IEC3D:}
The quantitative comparisons on IEC3D are presented in Table~\ref{tab:iec3d_results}. IEC3D contains real scanned industrial point clouds and mainly focuses on geometric defects. Our method achieves 89.93\% O-ROC and 97.74\% P-ROC, outperforming GMANet by 1.90 and 25.22 percentage points, respectively. The improvement in P-ROC is particularly significant, showing that our method can localize anomalous regions more accurately under real scanning conditions. This is important because real scanned point clouds often contain noise, non-uniform density, and partial observations. These results demonstrate that the proposed framework remains effective under real scanning noise, non-uniform density, and partial observations.

\begin{table}[t]
\centering
{\small
\begin{tabular*}{\columnwidth}{
@{\extracolsep{\fill}}cccc@{}
}
\toprule
Method & Pub./Year & O-ROC & P-ROC \\
\midrule
PC-FPFH & CVPR'22 & 56.80 & 58.00 \\
PC-MAE & CVPR'22 & 56.20 & 57.70 \\
BTF-Raw & CVPRW'23 & 49.30 & 55.00 \\
BTF-FPFH & CVPRW'23 & 52.80 & 62.80 \\
M3DM & CVPR'23 & 55.20 & 61.60 \\
Reg3D-AD & NeurIPS'23 & 57.20 & 66.80 \\
CPMF & PR'24 & 55.90 & 57.30 \\
IMRNet & CVPR'24 & 66.10 & 65.00 \\
R3D-AD & ECCV'24 & 74.90 & -- \\
ISMP & AAAI'25 & -- & 69.10 \\
PO3AD & CVPR'25 & 83.90 & 89.80 \\
MC3D-AD & IJCAI'25 & 84.20 & 75.90 \\
Point-Patch & CVPR'26 & 87.60 & -- \\
AF3AD & arXiv'26 & 91.50 & \underline{92.50} \\
SeDiR & CVPR'26 & \underline{93.30} & 81.00 \\
Ours & -- & \textbf{94.39} & \textbf{97.33} \\
\bottomrule
\end{tabular*}%
}
\caption{Quantitative results on Anomaly-ShapeNet. The results are reported as O-ROC\% and P-ROC\%. The best performance is in \textbf{bold}, and the second best is \underline{underlined}.}
\label{tab:anomaly_shapenet_results}
\end{table}

\begin{table}[t]
\centering
{\small
\setlength{\tabcolsep}{1mm}
\begin{tabular*}{\columnwidth}{
@{\extracolsep{\fill}}cccc@{}
}
\toprule
Method & Pub./Year & O-ROC & P-ROC \\
\midrule
BTF-Raw & CVPRW'23 & 68.93 & 70.17 \\
BTF-FPFH & CVPRW'23 & 48.05 & 53.40 \\
M3DM-PointMAE & CVPR'23 & 59.65 & 55.65 \\
M3DM-PointBERT & CVPR'23 & 58.38 & 54.17 \\
PC-FPFH & CVPR'22 & 86.99 & 64.26 \\
PC-FPFH+Raw & CVPR'22 & 87.05 & 66.86 \\
PC-PointMAE & CVPR'22 & 65.10 & 66.97 \\
Reg3D-AD & NeurIPS'23 & 75.18 & 69.89 \\
IMRNet & CVPR'24 & 75.05 & 71.64 \\
R3D-AD & ECCV'24 & 75.93 & 69.57 \\
GMANet & arXiv'25 & \underline{88.03} & \underline{72.52} \\
Ours & -- & \textbf{89.93} & \textbf{97.74} \\
\bottomrule
\end{tabular*}
}
\caption{Quantitative results on IEC3D. The results are reported as O-ROC\% and P-ROC\%. The best performance is in \textbf{bold}, and the second best is \underline{underlined}.}
\label{tab:iec3d_results}
\end{table}

\paragraph{Qualitative Results.}
Figure~\ref{fig:qualitative} shows qualitative anomaly localization results on ILGAD, Anomaly-ShapeNet, and IEC3D. The proposed method produces clear anomaly responses for logical defects in ILGAD, synthetic shape defects in Anomaly-ShapeNet, and real scanned defects in IEC3D.

\begin{table}[t]
\centering

\begin{tabular*}{\columnwidth}{@{\extracolsep{\fill}}lccccc@{}}
\toprule
Type & Geo. & Cov. & Rel. & O-ROC & P-ROC \\
\midrule
\multirow{4}{*}{Overall}
& \checkmark &  &  & 89.34 & 89.10 \\
& \checkmark & \checkmark &  & 90.79 & 96.16 \\
& \checkmark &  & \checkmark & 88.33 & 89.60 \\
& \checkmark & \checkmark & \checkmark & \textbf{91.90} & \textbf{96.28} \\
\midrule
\multirow{4}{*}{Logical}
& \checkmark &  &  & 90.47 & 90.28 \\
& \checkmark & \checkmark &  & 91.18 & 96.51 \\
& \checkmark &  & \checkmark & 90.99 & 90.66 \\
& \checkmark & \checkmark & \checkmark & \textbf{92.29} & \textbf{96.77} \\
\bottomrule
\end{tabular*}%

\caption{Ablation study of the proposed consistency branches on ILGAD.}
\label{tab:ablation_branch}
\end{table}

\begin{figure}[!t]
\centering
\includegraphics[width=0.95\columnwidth]{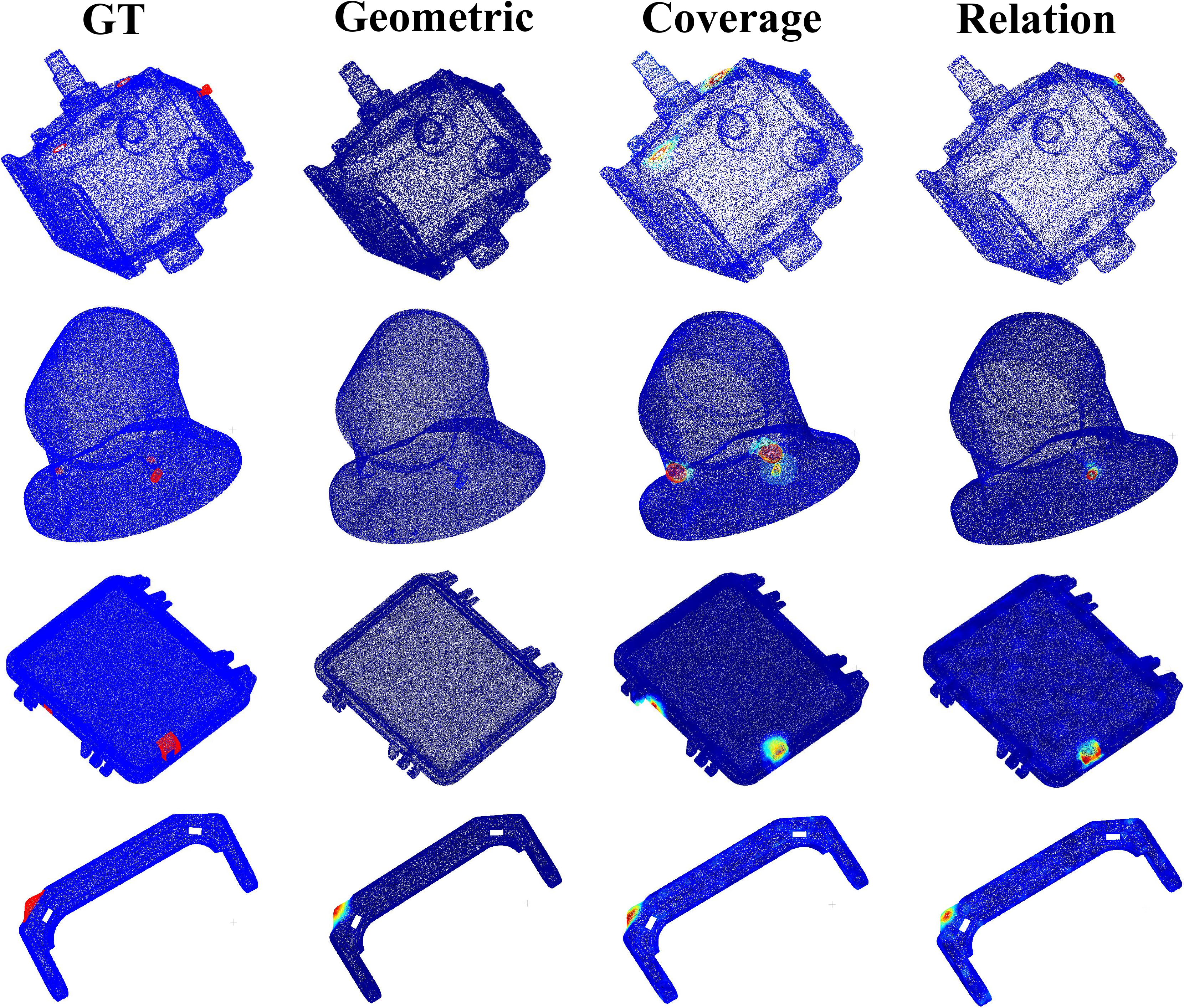}
\caption{Visualization of the three consistency responses on representative logical anomalies.}
\label{fig:branch_visualization}
\end{figure}
\subsection{Ablation Study}

Table~\ref{tab:ablation_branch} evaluates the complementarity of the three consistency branches. The geometric branch reaches 89.34\% O-ROC and 89.10\% P-ROC on the test set. Adding coverage branch increases P-ROC to 96.16\%, consistent with its role in localizing missing or unsupported structures. The relational branch provides modest but consistent gains on the subset containing at least one logical violation, where O-ROC/P-ROC improve from 90.47\%/90.28\% to 90.99\%/90.66\%. Combining all three components yields the best performance, reaching 91.90\%/96.28\% overall and 92.29\%/96.77\% on logical anomalies. Figure~\ref{fig:branch_visualization} provides a visual
interpretation of the three consistency branches. Coverage
consistency produces strong responses near missing or
insufficiently supported structures, while relational
consistency highlights regions with abnormal local
arrangements. Geometric consistency mainly responds to
visible local shape changes.

\section{Conclusion}
In this paper, we introduced ILGAD, a scalable 3D industrial benchmark dedicated to logical anomaly detection in point clouds. Unlike existing datasets that mainly focus on local geometric defects, ILGAD covers four types of logical defects, including existence errors, specification errors, pose errors, and assembly-state errors. We further proposed a consistency reasoning framework that evaluates a test point cloud through local geometric consistency, structure coverage consistency, and spatial relation consistency. These components jointly capture local structural inconsistency, missing normal structures, and abnormal neighborhood arrangements. Extensive experiments on ILGAD, Anomaly-ShapeNet, and IEC3D
demonstrate that our method achieves strong object-level detection
and point-level localization performance, showing the importance of
reasoning beyond local shape comparison for 3D logical anomaly detection.
Despite these promising results, the framework still relies on the normal template and accurate canonicalization. Severe missing structures, strong object symmetry, or noisy partial scans may affect reference-test matching. Future work will extend ILGAD to real scanned industrial scenes and explore more robust pose-equivariant matching for logical anomalies.

\bibliography{aaai2027}


\end{document}